\documentclass[letterpaper]{article} 
\usepackage{aaai25}  
\usepackage{times}  
\usepackage{helvet}  
\usepackage{courier}  
\usepackage[hyphens]{url}  
\usepackage{graphicx} 
\usepackage{natbib}  
\usepackage{caption} 
\usepackage{algorithm}
\usepackage{algpseudocode}
\usepackage{comment}
\usepackage{tabularx}
\usepackage{booktabs}
\usepackage{multirow}
\usepackage{mathtools, cuted}
\usepackage{amsmath}
\usepackage{newfloat}
\usepackage{listings}
\DeclareCaptionStyle{ruled}{labelfont=normalfont,labelsep=colon,strut=off} 
\floatstyle{ruled}
\newfloat{listing}{tb}{lst}{}
\floatname{listing}{Listing}
\title{RESQUE: Quantifying Estimator to Task and Distribution Shift for \\Sustainable Model Reusability}
\author{
    Vishwesh Sangarya and
    Jung-Eun Kim\thanks{Correspondence.}\\
}
\affiliations{
    Department of Computer Science, North Carolina State University\\

}

\usepackage{bibentry}

\begin{document}

\maketitle

\begin{abstract}
As a strategy for sustainability of deep learning, reusing an existing model by retraining it rather than training a new model from scratch is critical. In this paper, we propose REpresentation Shift QUantifying Estimator (RESQUE), a predictive quantifier to estimate the retraining cost of a model to distributional shifts or change of tasks. It provides a single concise index for an estimate of resources required for retraining the model. Through extensive experiments, we show that RESQUE has a strong correlation with various retraining measures. Our results validate that RESQUE is an effective indicator in terms of epochs, gradient norms, changes of parameter magnitude, energy, and carbon emissions. These measures align well with RESQUE for new tasks, multiple noise types, and varying noise intensities. As a result, RESQUE enables users to make informed decisions for retraining to different tasks/distribution shifts and determine the most cost-effective and sustainable option, allowing for the reuse of a model with a much smaller footprint in the environment. The code for this work is available here:\\ \url{https://github.com/JEKimLab/AAAI2025RESQUE}

\end{abstract}

%

\section{Introduction}
With the rapid and ever increasing use of deep learning models in everyday applications, a crucial aspect that is often overlooked is the adaptation of a model to changes and new data. Models need to be adaptive and capable of learning from these new inputs. This dynamic adaptation is essential for maintaining model performance and relevance over time. Particularly, the continuous series of changes a model may need to undergo subject to distributional shifts, or need for primary task updates. For example, new data with better relevant features may be introduced, the environment of deployment may undergo a change leading to distribution shift of the input data, or the prediction classes and task may change. Retraining helps the model remain effective with satisfactory performance for the desired deployed application while shortening the adaptation time. While new models can be developed from scratch when the data undergoes a distribution shift or the task changes, leveraging existing learned features is a resource-efficient option. 
Reusing models not only saves time and resources but also aligns with the principles of \emph{sustainable AI development}. This is of primary importance with the overwhelmingly increasing deployments of AI models in the current era, which in turn has given rise to the socially important field of Green AI \cite{schwartz2019green, MLSYS2022_462211f6,verdecchia2023systematic,10251541}. It is not uncommon for models to have several billion parameters \cite{pmlr-v202-dehghani23a, menghani2023efficient}, and as these models grow in complexity, their computational demands also increase. With the increased computational requirements, there is an accompanying need for resources required to perform the computation and byproducts of the computation to deal with. These demands place significant strain on available resources and highlight the importance of addressing both the direct and indirect impacts of AI model use. Consequently, there exists a need for sustainable and efficient development of AI models so as to reduce the environmental impact, especially when looking at the \emph{carbon footprint} \cite{10018371, strubell2019energy, bannour-etal-2021-evaluating,10550120} and energy consumption \cite{10.1145/3531146.3533234, strubell2019energy, 10018371, 
10550120} involved in the research, evolution, and utilization of AI models. 

With the goal of reusing models and reducing computation, energy consumption, and carbon emissions, we propose a novel estimating index called the REpresentation Shift QUantifying Estimator (RESQUE). This estimator provides a single index to predict the cost of retraining a model for new distributions or tasks before any computation is performed. 
\texttt{RESQUE} enables deep learning practitioners and researchers to quickly estimate the costs associated with adapting a model, offering valuable insights for informed decision-making. This quantifiable estimator helps them achieve sustainability goals when using and developing the models. For the context of distribution shifts, the estimator referred to as \texttt{RESQUE}\textsubscript{dist}, measures the shift in the model's representation outputs between the original and new distribution. To obtain \texttt{RESQUE}\textsubscript{dist}, the input data is propagated through the model just \emph{once} with \emph{no backward} propagation or computation. 
For the case of a new target task, the estimator, specifically termed as \texttt{RESQUE}\textsubscript{task}, quantifies the separation in class decision boundary of the new target task in the representation space, using the original task model. 
\texttt{RESQUE} obtained after the forward pass serves as an index to estimate the resources required to adapt the model to a new target task or new distribution. Through extensive experiments and by theoretical reasoning, we show that a lower value of \texttt{RESQUE} correlates with a lower cost of retraining to a new target task or new distribution.

Additionally, we show that \texttt{RESQUE} not only correlates with important sustainability costs such as energy consumption and carbon emissions, but also has strong correlation with computational related measures such as training epochs, gradients, and model parameter change magnitudes. Furthermore, \texttt{RESQUE} is model/architecture-agnostic, proving effective against either convolutional networks or transformers. Our experiments span a wide range of target tasks, various vision datasets, and different types and intensities of distribution shifts, demonstrating the practical value of \texttt{RESQUE}. 
Using \texttt{RESQUE} aids in meeting resource and sustainability targets while reducing the computational effort required for adapting AI models.

\section{Related Work}
Sustainable computing and Green AI
\cite{schwartz2019green, verdecchia2023systematic,MLSYS2022_462211f6, 10251541} are gaining more and more attention due to their potential impacts on the modern AI era of large models. As the demand for AI grows \cite{liu2024empirical, 10550120}, the need for efficient and sustainable computing practices has become important, leading to an important research domain. \cite{verdecchia2023systematic, 10550120} perform a detailed study regarding the potential quantification of undesirable effects of AI, they highlight the importance of reporting energy consumption and carbon emissions as key sustainability measures. Other studies \cite{MLSYS2022_462211f6, schwartz2019green, 10251541} bring forth the need for a sustainable and greener development of AI, while highlighting the environmental consequences of neglecting the impact of inconsiderate use of resources.

\cite{10.1145/3510003.3510221, strubell2019energy} conduct detailed studies regarding energy usage of neural networks. In particular \cite{strubell2019energy, bannour-etal-2021-evaluating, 10.1145/3510003.3510221} explore the increasing energy demands of neural networks from the perspective of computing power, operational costs, and training time. Along with energy consumption, \cite{JMLR:v21:20-312, patterson2021carbon} highlight the absence of carbon emissions reporting associated with deep learning research. Besides these studies, other works such as \cite{xu2023energy, GARCIAMARTIN201975, mcdonald-etal-2022-great} discuss the environmental impact of training models from a carbon emissions perspective. To aid in tracking carbon emissions and energy usage associated with deep learning development, \cite{schmidt2021codecarbon, anthony2020carbontracker, lacoste2019quantifying} propose useful quantifying strategies, including frameworks and libraries which can accurately log all energy consumption and resulting carbon emissions from training. \cite{10.1145/3531146.3533234} demonstrate that training Vision Transformers \cite{dosovitskiy2021an} results in significantly higher carbon emissions than convolutional networks due to their greater energy requirements from extensive training characteristics. On a similar stream, \cite{strubell2019energy} assess the substantial energy consumption of transformer-based models.

Changes in data due to varying input distributions can arise due to various factors, as elaborated in \cite{hendrycks2019benchmarking, arjovsky2020invariant, Hendrycks_2021_ICCV}. To adapt to the specific distributional changes, augmentation strategies \cite{hendrycks2020augmix, liu2022randommix, zhang2018mixup, kimICML20, lee2020smoothmix} were effective to a limited extent. Although these techniques improve performance compared to a standard trained model against certain types of shifts, they are computationally costly and require training models from scratch. Moreover, the performance improvements are for a limited number of shifts in the distributions, while providing no benefits or a drop in performance for other types of distribution shifts. Studies such as \cite{geirhos2020generalisation, yin2020fourier, ford2019adversarial} detail the non-uniform performance gains, where improvements in performance cannot be controlled and may result in lower than acceptable performance at the end of training. Adaptation during test time, such as \cite{lim2023ttn, niu2022efficient, goyal2022testtime, wang2022continual}, addresses the computational cost concerns by having additional tuned components that can be added to the model. However, these approaches heavily rely on batch data and tend to provide uncertain and low-confidence outputs when there is a greater variation of distribution shifts in the input.

Gradients play a crucial role in evaluating the complexity and difficulty of learning when training models, as demonstrated in \cite{agarwal2022estimating, lee2020gradients, huang2021on}. These studies tell us that difficult samples produce greater gradients and thus require larger computation costs for convergence. 
\cite{Sangarya2024AIES,Sangarya2023NeurIPSW} provide measures for evaluating distribution shifts when a shifted input sample is only derived from a corresponding original sample, which is not a realistic scenario. 
In particular, \cite{Sangarya2023NeurIPSW} only operates on individual shifts without the ability to evaluate multiple types of shifts simultaneously. Moreover, both studies only focused on shifts in data but did not take into account the practical scenario where the task itself undergoes a change. Additionally, studies such as 
\cite{stacke2020measuring} measure changes in layer-wise representations to analyze different shifts in pathology data.

Adjusted Rand Index \cite{hubert1985comparing}, is a useful metric for comparing clustering algorithms. However, it can also be used to assess the performance of supervised classification and feature selection as demonstrated by \cite{santos2009use}. With regard to the case of task change, we refer to research in various subfields of deep learning, such as \cite{ravi2016optimization, oreshkin2018tadam, caruana1997multitask}, which defines a new task as any change in the class data or the introduction of completely new classes of associated data. \cite{achille2019task2vec} is designed to measure task similarity and perform an evaluation of pretrained models. However, adapting to a new task from a current task model requires dataset and task-specific analysis to estimate retraining costs. Likewise, although \cite{li2023guided} consider factors such as dataset and model characteristics, they do not specifically address the magnitude of retraining needed to adapt a model to a new task.

\section{RESQUE}
In this section, we provide details of \texttt{RESQUE}\textsubscript{dist} for distributional shifts and \texttt{RESQUE}\textsubscript{task} for change of task.

\subsection{Distribution Shift}
\label{RESQUE:distribution shift}
In this scenario, we consider the case when a model needs to be updated and adapted as it saw a distributional shift in data - for example, because some ground truths might be changed, some data samples might become stale, or some new data samples might need to come into play, etc. We explain how to obtain and use \texttt{RESQUE}\textsubscript{dist}, which is a predictive quantifier for model retraining to distributional shifts. It quantifies the distance between representations of the original and the shifted distribution so as to determine how \emph{familiar} or \emph{foreign} the distribution is.

\subsubsection{Normalized Embedding Vectors for Distributions}
For a realistic scenario where a distribution-shifted sample has no corresponding clean image, or the number of distributed shifted samples may not be equal to the number of original clean samples, we make use of all the samples to generate representation embedding vectors. For the case of distribution shifts, since there is no change in the original task and therefore no change in the number of classes, we obtain the class-wise representation embedding vectors for each class individually. To obtain the data representations, we use the output representation from the final convolutional layer in convolutional layers and the final dense layer in Vision Transformers. We perform a forward pass of each dataset and obtain a summed embedding representation vector of each class as follows,
\begin{equation}
    V^{D}_{l, sum} = \sum_{i=0}^{n_l} {V^{D}_{l, i}}
    \label{eq:RESQUE_representation_sum}
\end{equation}
where, $V^{D}_{l, i}$ represents the flattened representation of sample $i$ with class label $l$ within dataset $D$. For each class $l$, where $n_l$ is the number of samples within the class, the representation vectors are summed to obtain the summed embedding representation vector $V^{D}_{l, sum}$.
The summed embedding vector for each class label $l$ is then normalized as follow,
\begin{equation}
    V^{D}_{l, norm} = \frac{V^{D}_{l, sum}} {|| V^{D}_{l, sum} ||}
    \label{eq:RESQUE_norm}
\end{equation}

\subsubsection{RESQUE\textsubscript{dist} from Normalized Embeddings}
The final representation angle is obtained by averaging the inverse cosine angle between each class' normed representation vector for the original dataset and the distribution-shifted dataset. Let $O$ represent the original dataset, $S$ represent the distribution shifted dataset, and $k$ represent the number of classes. \texttt{RESQUE}\textsubscript{dist} is then obtained as,
\begin{equation}
 RESQUE_{dist} = \frac{
 \sum_{i=0}^{k} \arccos(V^{O}_{i, norm}, V^{S}_{i, norm})
 }{k}
 \label{eq:RESQUE_dist}
\end{equation}

\begin{algorithm}
\caption{Initialize cluster representation centroids}
\textbf{Input:} Number of samples $n_s$, \\        
\phantom{Input: } Number of labels $k$,\\       \phantom{Input: } Model $M$ producing representation output ($R$, $Y$), \\
\textbf{Output:} Centroids matrix $CentM$
\begin{algorithmic}
\State $CentM \gets \textit{Empty Vector of size k}$
\For{$i \gets 1$ to $k$}
    \State $current\_centroid \gets \overrightarrow{0}$
    \State $label\_count \gets {0}$
    \For{$j \gets 1$ to $n_s$}
        \State $(R, Y) \gets M(X_j)$ \algorithmiccomment{$X_j$ is the $j$th sample}
        \If{$Y == i$}
        \State $current\_centroid \gets current\_centroid + R$
        \State $label\_count \gets label\_count + 1$
        \EndIf
    \EndFor
    \State $CentM_i \gets current\_centroid/label\_count$
\EndFor
\end{algorithmic}
\label{alg:algorithm_centroid}
\end{algorithm}

\subsection{Change of Task}
\label{RESQUE:change_of_task}
For the first, we consider a case in which a model is required to be updated for application toward a new target task. We show how a model can be retrained and reused to meet such a requirement. To characterize the resources and cost of retraining to a new task, we make use of the separation of classes within the latent space of representations. The separation of classes in the representation latent space is an effective gauge of the model's ability to provide accurate outputs. We relate this to the distance between decision boundaries for classes within the representation space, where distinct boundaries result in correct and confident predictions, whereas class boundaries that overlap or are close to each other lead to low confidence and incorrect predictions. We use this trait of deep learning models to evaluate how well a model's current learned features and representation space translate into learning a new task.

\subsubsection{Quantifying Class Separation from Known Task Data to New Task Data}
We use the representation of the data produced by the model and a clustering algorithm to assign cluster labels for each data sample in the representation latent space. Using the assigned cluster labels, we derive our estimator, \texttt{RESQUE}\textsubscript{task}, by applying the Adjusted Rand Index \cite{hubert1985comparing}. \texttt{RESQUE}\textsubscript{task} can quantify the effectiveness of class separation by using the cluster labels and the true data labels. To obtain the cluster labels, we first retrain the original model with the new task just for a \emph{single} epoch. This creates a representation that is primarily of the original task but incorporates features of the new task and updates the output classes of the model to match the new task classes. Following this step, the new task data is used to perform a forward pass and assign a label, which is a representation label, obtained via clustering on all samples. 

\subsubsection{Clustering Mechanism for the New Unknown Task}
The data representation is obtained from the final convolution layer for a convolutional network' case while from the final dense layer in the last transformer encoder block for a vision transformer's case. To obtain the representation labels by clustering, we use KMeans with 3 different centroid initialization techniques as follows: 
\begin{enumerate}
  \item[1.] Using the original data to obtain centroids as detailed in Algorithm~\ref{alg:algorithm_centroid},
  \item[2.] Initializing by Kmeans++ \cite{arthur2007k}, and
  \item[3.] Initializing by random cluster assignment, and selecting the cluster with least entropy among 20 random initialization seeds.
\end{enumerate}
The above three initialization schemes result in similar clustering labels. We use Algorithm~\ref{alg:algorithm_centroid} due to its computational efficiency as it does not require random re-initialization or iterative assignment of centroids.

In Algorithm~\ref{alg:algorithm_centroid}, we begin by creating an empty vector that has a size equal to the number of class labels, as depicted in line 1. $CentM$ is a vector of size $k$ when initialized, but as the classes' centroids are obtained, each entry in $CentM$ is a vector itself. In the end, $CentM$ is a matrix of size $k$ x size of the flattened representation.

\subsubsection{Adjusted Rand Index to Quantify Class Separation}
Adjusted Rand Index takes the value $0$ for purely random clustering, and $1$ for identical clustering. For our estimator, it is required to have a low value for decision boundaries which are well separated and high value for boundaries which overlap and result in incorrect representation cluster labels. Hence, for our estimator, \texttt{RESQUE}\textsubscript{task}, we take the complement of Adjusted Rand Index and define it in Eq.~\eqref{eq:RESQUE_task}. In Eq.~\eqref{eq:RESQUE_task}, $tl$ represents the total count of true labels for each label in the contingency table of true labels vs. representation labels via clustering. $rc$ represents the summed values of representation cluster labels in the contingency table. A contingency table in this scenario is a matrix that summarizes the number of samples belonging to the same cluster or having the same label in both clustering scenarios. Here, by `both clustering scenarios,' we refer to the representation-based clustering labels and the true labels. $n_c$ is the number of cluster labels, which is equal to the number of class labels. $n_{i,j}$ represents the value in each entry of the contingency table, which is common to both cluster labels for a given label $i$ and $j$. $n_s$ is the total number of samples. 

The overall formulation \texttt{RESQUE}\textsubscript{task} is as follows,
\begin{equation}
\begin{aligned}
&    RESQUE_{task} = \\
&    \frac{\frac{1}{2} \Big[\sum_{i=0}^{n_c}\hspace{-0.1cm} {rc_{i}\choose{2}} \hspace{-0.1cm}+ \hspace{-0.1cm}\sum_{j=0}^{n_c} \hspace{-0.1cm}{tl_{j}\choose{2}}\Big] - \sum_{i,j=0}^{n_c} \hspace{-0.1cm}{n_{i,j}\choose{2}}}{\frac{1}{2} \Big[\sum_{i=0}^{n_c}\hspace{-0.1cm} {rc_{i}\choose{2}} \hspace{-0.1cm}+\hspace{-0.1cm} \sum_{j=0}^{n_c} \hspace{-0.1cm}{tl_{j}\choose{2}}\Big] \hspace{-0.1cm} - \hspace{-0.1cm} \Big[\sum_{i=0}^{n_c}\hspace{-0.1cm} {rc_{i}\choose{2}} \sum_{j=0}^{n_c} \hspace{-0.1cm}{tl_{j}\choose{2}}\Big]/{n\choose{2}}}
    \label{eq:RESQUE_task}
\end{aligned}
\end{equation}

\begin{table}[t]
    \footnotesize
    \centering
    \caption{Correlation coefficients (and associated p-value) for multiple models}
    \vspace{-0.3cm}
    \resizebox{1.0\linewidth}{!}{
    \begin{tabular}{llrrr}
    \toprule
    Model & Correlation & Epochs & GradNorm & Param. change \\
    \midrule
    ResNet18 & Pearson & 0.77 (6e-07) & 0.74 (2e-06) & 0.70 (1e-05) \\
    (CIFAR100) & Spearman & 0.97 (1e-19) & 0.97 (1e-20) & 0.95 (6e-17) \\
    \hline
    ViT & Pearson & 0.87 (2e-10) & 0.88 (1e-10) & 0.84 (3e-09) \\
    (CIFAR10) & Spearman & 0.88 (9e-11) & 0.88 (8e-11) & 0.88 (1e-10) \\
    \hline
    VGG16 & Pearson & 0.93 (2e-08) & 0.95 (1e-16) & 0.94 (1e-15) \\
    (SVHN) & Spearman & 0.92 (5e-13) & 0.95 (9e-16) & 0.93 (3e-14) \\
    \bottomrule
    \end{tabular}
    }
    \label{tab:noise_correlation}
    \vspace{-0.2cm}
\end{table}

\begin{figure*}[hpt]
    \centering
    {\includegraphics[width=0.85\linewidth]{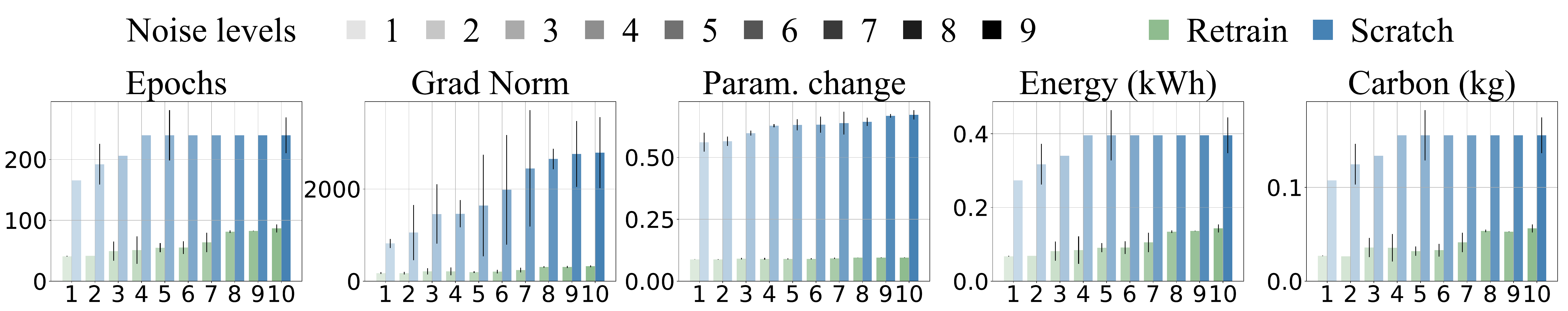}}
    \vspace{-0.2cm}
    \caption{Retraining vs. Training from scratch on the SVHN dataset with Gaussian noise using VGG16}
    \vspace{-0.25cm}
    \label{fig:DistributionShift_ScratchVsRetrain}
\end{figure*}

\section{Retraining Measures}
Retraining measures refer to the resources and costs expended when a model is adapted to a new task or distribution. We demonstrate that \texttt{RESQUE} is strongly correlated with these measures and serves as an effective estimator. Using \texttt{RESQUE}, users can estimate the resource expenditure needed for retraining the models. We evaluate several measures including carbon emissions, energy consumption, epochs, total gradient norm, and normalized parameter change.

\subsubsection{Carbon Emissions}
As highlighted in earlier sections, carbon emissions reporting \cite{JMLR:v21:20-312} is vital for sustained development as it represents the environmental cost linked to model training. We utilize a carbon tracking library \cite{schmidt2021codecarbon}, to track the carbon emissions associated with training. The library estimates the carbon footprint by taking into account the method of energy generation in the region where the GPUs are used for model training, and, along with the energy consumed, calculates the estimated carbon compound byproducts.

\subsubsection{Energy Consumption}
Energy consumption is another vital sustainability cost that needs to be measured and reported. We utilize the same library \cite{schmidt2021codecarbon} used for carbon tracking to measure the energy consumed for the entire training and retraining process. To measure the energy consumed, the energy of processes that operate on the GPU, CPU, and RAM are taken into account to provide the total energy consumption.

\subsubsection{Epochs}
Along with carbon emissions and energy consumption, the training cost can be quantified by the number of epochs required to reach the desired performance when adapting to new distributions or tasks. We provide detailed empirical results showing that \texttt{RESQUE} consistently aligns with the number of epochs, confirming its effectiveness as an estimator. The epochs reported are the number of epochs a model requires to reach a predefined test accuracy. Due to common hyperparameters, as additional termination conditions, training is halted after 25 or 50 epochs if the accuracy falls within 0.5\% or 1\% of the selected cutoff, respectively.

\subsubsection{Total Gradient Norm}
Gradient norm is a common measure of the difficulty of learning algorithms. Research studies such as \cite{agarwal2022estimating, huang2021on, lee2020gradients} use gradient norm as a proxy for sample difficulty. In this paper, we report the gradient norm by aggregating the gradient norm at each time interval during retraining. The final gradient norm value represents the total magnitude of gradients the model experienced from the beginning to the end of the training.

\subsubsection{Normalized Parameter Change}
The change in the model parameters denotes the extent of update a model is supposed to undergo when estimating the retraining measure. 
A larger normalized parameter change indicates significant model deviation and higher resource use, conversely, a lower change suggests lower costs and faster convergence.

Similar to \cite{zhang2022layers}, we calculate parameter change but instead of comparing initial and final values, we aggregate changes between consecutive time steps of training and retraining. For parameters of layer $l$ at time $t$, the normalized change $N_{l,t}$, between current parameters and the previous time step's parameters is given by
\begin{equation}
    N_{l, t} = \left \lVert W_{l,t} - W_{l,t-1}  \right \rVert_2 /\sqrt{\left \lVert W_{l,t}  \right \rVert}
\end{equation}
Here, $t$ $(\geq 1)$ represents the current time interval, while $t - 1$ represents the previous time interval. The distance between current parameters $W_{l, t}$ and the parameters from the previous time interval $W_{l, t-1}$ is calculated for each layer in the model. The final value of the normalized parameter change is obtained by summing up the total change per time interval, per layer, and averaged by the number of layers.
\begin{figure*}[ht]
    \centering
    {\includegraphics[width=\linewidth]{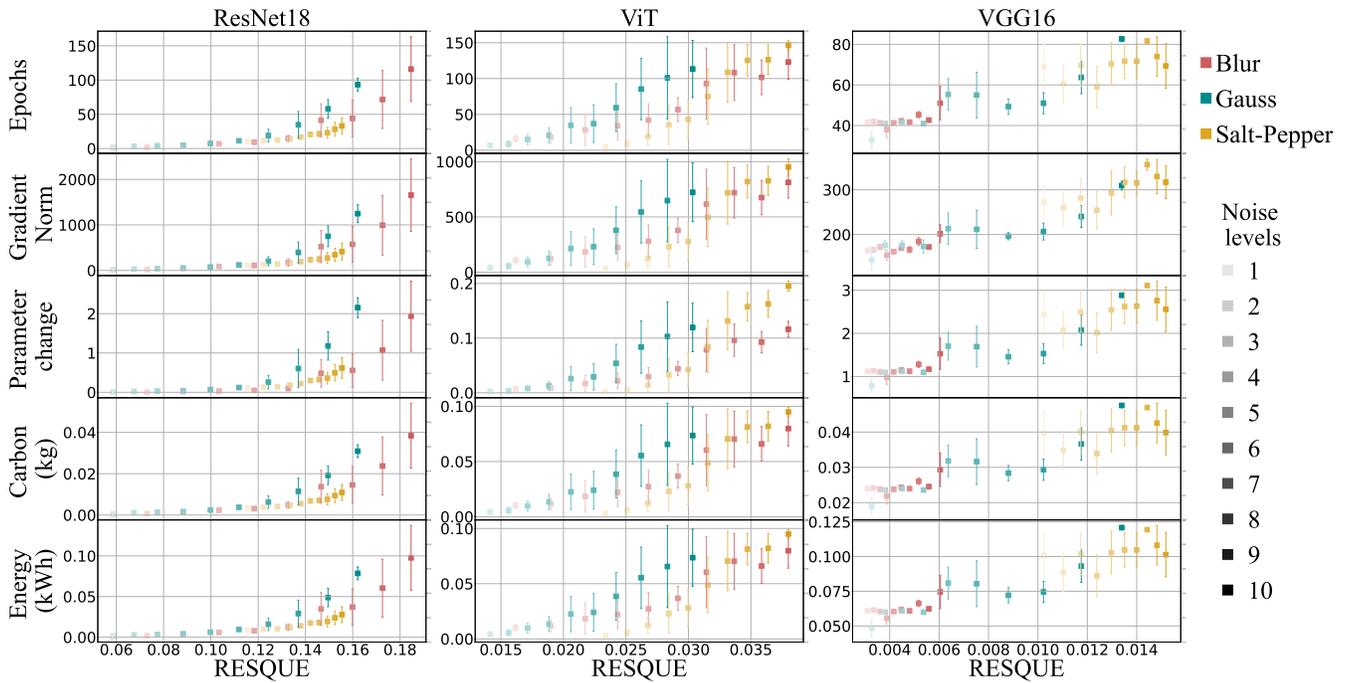}}
    \vspace{-0.5cm}
    \caption{\texttt{RESQUE} and retraining measures for different models and datasets
    }
    \label{fig:DistributionShift_AllResults}
    \vspace{-0.25cm}
\end{figure*}

\section{Experiments}
In this section, we discuss the experiment details and results for distribution shifts and task changes. All experiment results are an average of three runs. More details regarding hyperparameters, hardware configurations and software setups are provided in the Appendix.

\subsection{Distribution Shift}
In this section, we assess \texttt{RESQUE}\textsubscript{dist} in the context of distributional shifts using three datasets : CIFAR10 \cite{krizhevsky2009learning}, CIFAR100 \cite{krizhevsky2009learning}, and SVHN \cite{netzer2011reading}. We utilize 3 types of noises - Gaussian noise, Image Blur, and Salt-Pepper noise, which represent various real world noises that can occur due to hardware changes, environmental factors, or artifacts in the data, respectively. For each noise type, we generate 10 levels of noise intensity, with level 1 corresponding to minimal noise and level 10 aligning with severity 4 as described in \cite{hendrycks2019benchmarking}. To represent the reusability of pre-existing models, we initially train a randomly initialized model on the original data distribution until it achieves the minimum required accuracy for each dataset for fair comparisons. While tuning hyper-parameters for higher noise levels could expedite convergence, it may hinder comparisons across different noise types and levels. Therefore, we maintain consistent hyper-parameters across all experiments for consistency. 

It is important to note that, for the initial model training on original data without distributional shifts, we utilize 70\% of the entire dataset. For retraining to data with distributional shifts, we utilize 50\% of the dataset with added noise. The 20\% excess is the overlap data that was common in the original distribution and the new distribution shifted data.

Fig.~\ref{fig:DistributionShift_ScratchVsRetrain} illustrates the retraining measures for VGG16 trained from scratch versus retrained on SVHN under varying intensities of Gaussian noise. The results clearly show that retraining a model requires significantly fewer resources than training from scratch. All retraining measures for retraining are substantially lower compared to those for training a new model, highlighting the efficiency of retraining to distribution shifts.

For evaluating \texttt{RESQUE}\textsubscript{dist} as an estimator for the retraining measures, we evaluate convolutional networks and vision transformers, retrained to various noise types and levels on different datasets. We train and retrain ResNet18 \cite{he2016deep} on CIFAR100, VGG16 \cite{simonyan2014very} on SVHN, and ViT \cite{dosovitskiy2021an} on CIFAR10. For the Vision Transformer, we utilize a ViT model with a patch size of 4, comprising 8 transformer blocks, a latent vector size of 512, 8 attention heads, and an MLP with a hidden layer size of 1024.

Figure~\ref{fig:DistributionShift_AllResults} displays the correlation between \texttt{RESQUE}\textsubscript{dist} and the retraining measures, for all three models on the three datasets. Across various datasets and architectures, \texttt{RESQUE}\textsubscript{dist} consistently aligns with the retraining measures for different types of distribution shifts. Furthermore, Table~\ref{tab:noise_correlation} provides the Pearson and Spearman correlation coefficients, along with associated p-values, between \texttt{RESQUE} and all the retraining measures, demonstrating a strong correlation between \texttt{RESQUE}\textsubscript{dist} and the retraining measures.

\begin{figure}[ht]
    \centering
    {\includegraphics[width=0.95\linewidth]{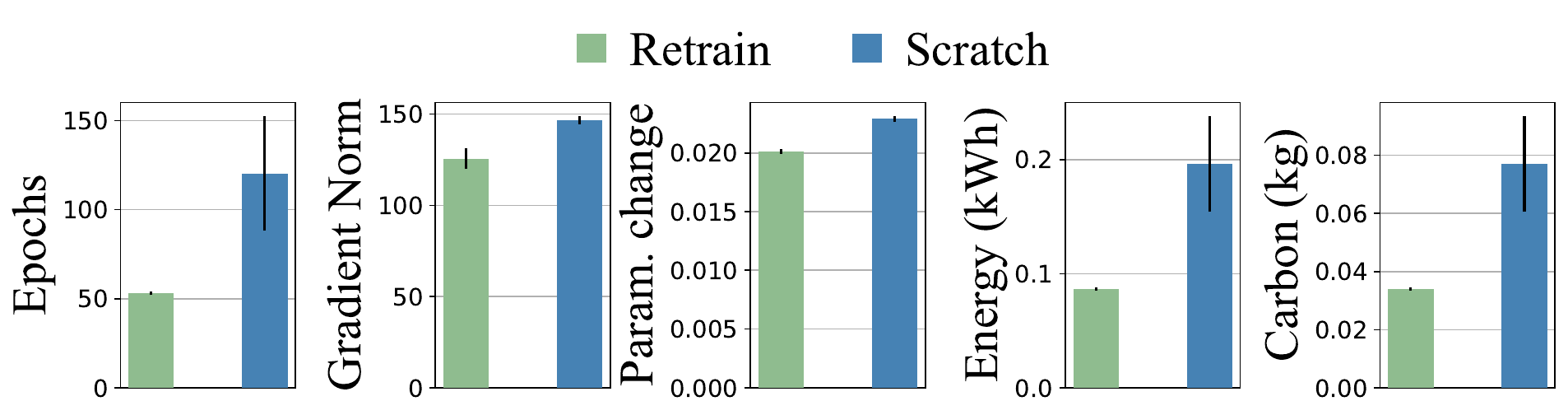}}
    \vspace{-0.1cm}
    \caption{As for the new target task, CIFAR10, comparisons of retraining from Food101 vs. training from scratch. Retraining consumes significantly less resources, epochs, energy, and carbon, than training from scratch.}
    \vspace{-0.5cm}
    \label{fig:NewTask_ScratchVsRetrain}
\end{figure}

\begin{table}[t]
    \centering
        \caption{Correlation coefficients (and p-values) of ResNet and ViT for different new target tasks.}
    \vspace{-0.3cm}
    \resizebox{1.0\linewidth}{!}{
    \begin{tabular}{lllll}
    \toprule
    Model & Correlation & Epochs & Grad Norm & Param. change \\
    \midrule
    ResNet18 & Pearson & 0.86 (0.012) & 0.82 (0.022) & 0.65 (0.100) \\
    (CIFAR10) & Spearman & 0.96 (4e-04) & 0.92 (0.002) & 0.75 (0.052) \\
    \hline
    ResNet18 & Pearson & 0.75 (0.050) & 0.74 (0.053) & 0.71 (0.071) \\
    (CIFAR100) & Spearman & 0.89 (0.006) & 0.85 (0.014) & 0.78 (0.036) \\
    \hline
    ResNet18 & Pearson & 0.93 (0.002) & 0.83 (0.020) & 0.93 (0.002) \\
    (GTSRB) & Spearman & 0.75 (0.052) & 0.75 (0.052) & 0.64 (0.119) \\
    \hline
    ViT B/16 & Pearson & 0.93 (7e-04) & 0.93 (7e-04) & 0.92 (0.001) \\
    (CIFAR10) & Spearman & 0.83 (0.009) & 0.85 (0.006) & 0.71 (0.046) \\
    \bottomrule
    \end{tabular}
    }
    \vspace{-0.25cm}
    \label{tab:task_correlation}
\end{table}

\subsection{Task Change}
\label{sec:experiments_task_change}
We evaluate the effectiveness of \texttt{RESQUE} to estimate resources for learning a new task from an original task. First, we compare training a model to a target task from scratch vs. retraining a model from the original task. For training from scratch, we randomly initialize a new model and train it on the target task. We set a common cutoff accuracy for a fair comparison. Fig.~\ref{fig:NewTask_ScratchVsRetrain} displays the cost indicators for ResNet18, which was trained on Food101 for the cases of retraining vs. training just from scratch. The target task is CIFAR10. It is clear that for all retraining measures, retraining a model requires significantly fewer resources than training a model from scratch. When retraining an original task model to a new task, since the features of the new task are not known, the model experiences higher initial gradient norm and parameter change. However, this is not the case when retraining to distribution shifts, since the model has already learned a good amount of features of the data, resulting in lower gradient norm and parameter changes, as shown in Fig.~\ref{fig:DistributionShift_ScratchVsRetrain}.

\begin{figure*}[ht]
    \centering
    {\includegraphics[width=0.95\linewidth]{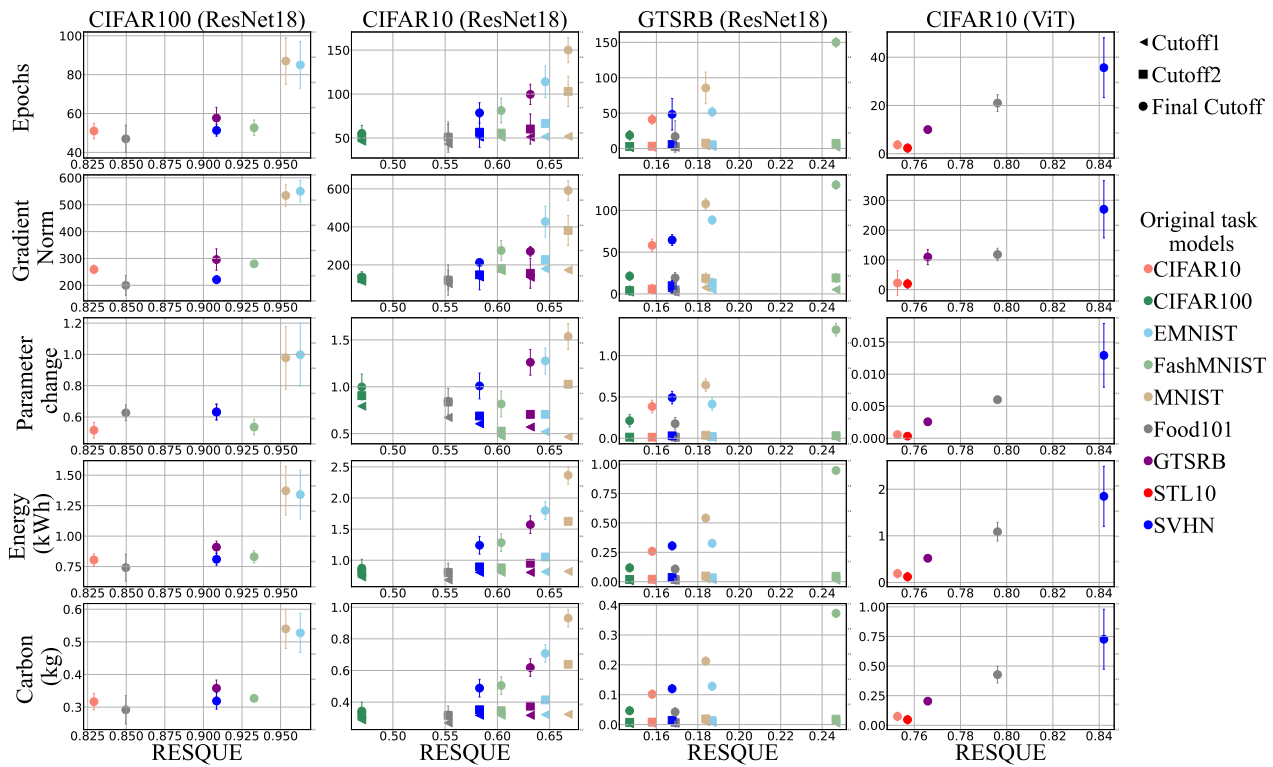}}
    \caption{RESQUE vs. resource measures for ResNet and ViT retrained to different new target tasks. A positive relation between RESQUE and resource measures is exhibited.}
    \label{fig:tasks_all}
    \vspace{-0.25cm}
\end{figure*}

\begin{figure}[ht]
    \centering
    \vspace{-2pt}
    {\includegraphics[width=0.8\linewidth]{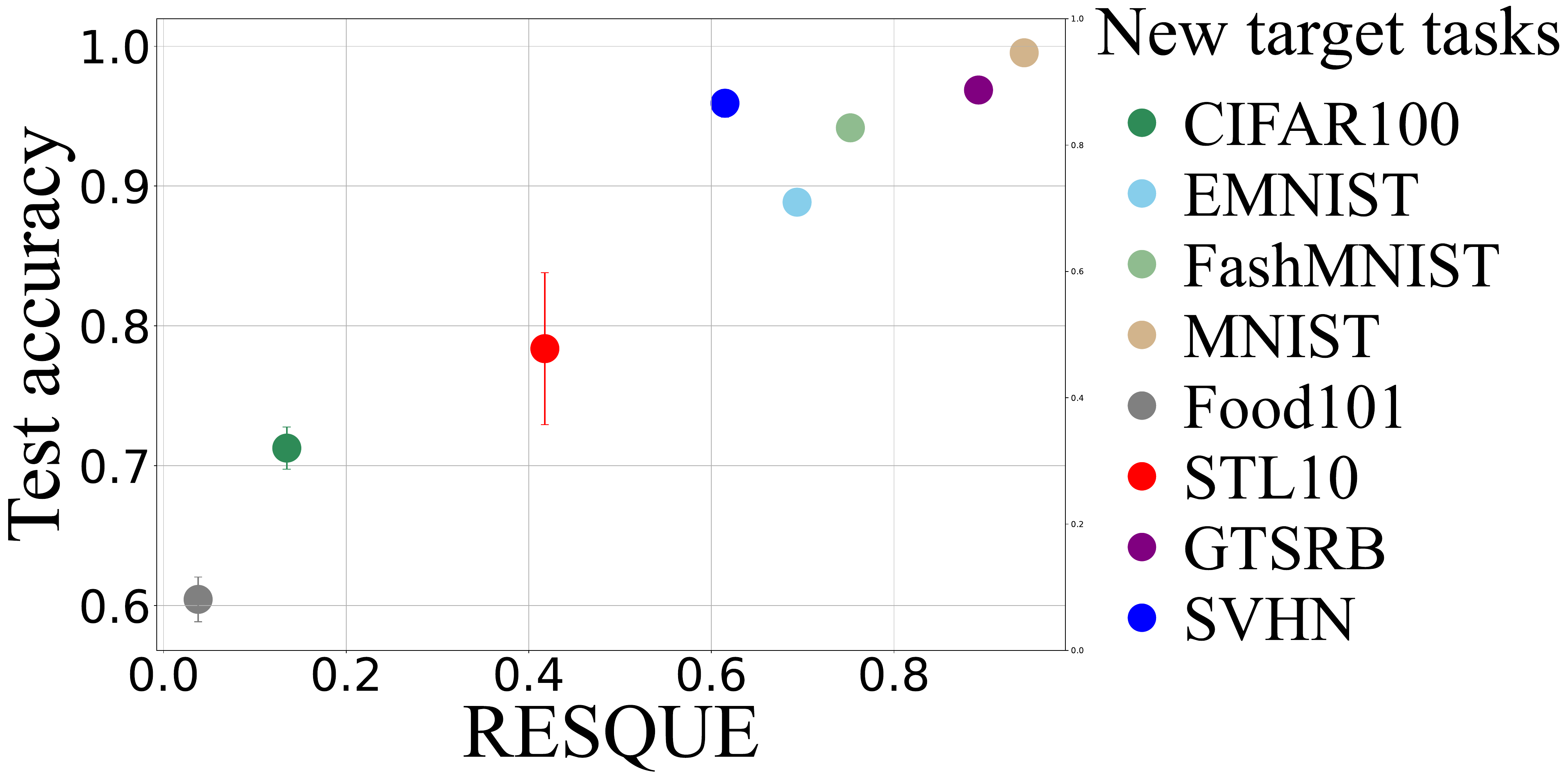}}
    \vspace{-0.25cm}
    \caption{CIFAR10 model trained to different target tasks. Error bars represent deviations. RESQUE has a strong linear relation with the peak performance a model can achieve on new tasks.}
    \label{fig:Task_CIFAR10_to_rest}    
    \vspace{-0.5cm}
\end{figure}

\subsubsection{Experiments Across Different Original Tasks}
\label{sec:task_change_different_original_tasks}
We show how well \texttt{RESQUE}\textsubscript{task} and the retraining measures are aligned for different original tasks. The original tasks evaluated in Fig.~\ref{fig:tasks_all} are trained on CIFAR10 \cite{krizhevsky2009learning}, CIFAR100 \cite{krizhevsky2009learning}, EMNIST \cite{cohen2017emnist}, Fashion MNIST \cite{xiao2017fashion}, Food101 \cite{bossard2014food}, GTSRB \cite{stallkamp2012man}, MNIST \cite{lecun1998gradient}, and SVHN \cite{netzer2011reading}. We use ResNet18 and ViT-B/16 \cite{dosovitskiy2021an}, and perform clustering to obtain \texttt{RESQUE}\textsubscript{task} using the last convolutional layer of ResNet18, and the final dense layer in the last transformer encoder block of the ViT. During training and retraining, images are resized to 224x224 for ViT-B/16 and to 32x32 for ResNet18. To achieve better performance and to ensure consistent training conditions across all new target tasks, we retrain all layers of the model, which provides better accuracy performance than retraining only the final few layers as outlined in \cite{deng2023towards}. 

Fig.~\ref{fig:tasks_all} depicts the relation between \texttt{RESQUE}\textsubscript{task} and all the retraining measures for ResNet and ViT on different target tasks. For the target tasks, CIFAR10 and GTSRB, we use multiple cutoff accuracies to evaluate how learning differs from the early stage to the later stage. From the figure, it is evident that \texttt{RESQUE}\textsubscript{task} aligns with the retraining measures and has a strong correlation.

A lower value of \texttt{RESQUE}\textsubscript{task} indicates that the model will require fewer resources to converge. For the target tasks, CIFAR10 and GTSRB, during the early learning stage, the trends exhibited are linear for all the cases of the original tasks. However, as training progresses, the number of epochs required by models with higher \texttt{RESQUE}\textsubscript{task} increases substantially. We relate this to the difficulty of learning and the usefulness of task and feature similarity. 

We see that, in an original task model with aligned and well-learned initial features, the increase in epochs from low accuracy to high cutoff accuracy is smaller. As opposed to that, for a model with poorly learned initial features, there is a substantial increase in epochs from low accuracy to high cutoff accuracy. 
 
This is due to the fact that learning becomes progressively more difficult for models with less aligned or poorly learned features. This implies two crucial learning challenges - not only does the model lack relevant prior knowledge, but additionally, it also struggles to extract effective features from its initial training, leading to steeper learning curves and thus resulting in a larger number of epochs and resources expended. For ViT B/16 in Fig.~\ref{fig:tasks_all}, some of the prior task models fail to reach a high cutoff accuracy. CNNs overfit and achieved low accuracy on STL10 due to the limited sample size and the need for resizing to 32x32, hence, we did not use STL10-trained CNNs for new target task retraining.

Table~\ref{tab:task_correlation} provides numerical correlation values as evidence of a strong correlation between \texttt{RESQUE}\textsubscript{task} and the retraining measures. \texttt{RESQUE}\textsubscript{task} has a strong positive Pearson correlation and Spearman correlation coefficient, with a low p-value for both experiments. This indicates that there is a strong and statistically significant relationship between \texttt{RESQUE}\textsubscript{task} and the retraining measures.

\subsubsection{Experiments Across Different Target Tasks}
\label{sec:task_change_different_target_tasks}

\texttt{RESQUE}\textsubscript{task} can also accurately estimate the performance an original task model can achieve on different new target tasks. This can help profile and categorize task similarity and provide useful information for retraining a current model in future instances. For retraining to different target tasks, we retrain the original task model for a fixed number of epochs across all new target tasks. Fig.~\ref{fig:Task_CIFAR10_to_rest} illustrates the relation between \texttt{RESQUE}\textsubscript{task} and the highest test accuracy reached on each of the new task datasets (original task is CIFAR10). Using \texttt{RESQUE}\textsubscript{task}, we obtain an accurate estimation regarding the peak performance a model can attain on the new task. 
\section{Conclusion}
We introduced a novel metric, \texttt{RESQUE}, to estimate the various resources that would be expended when reusing a model by adapting to distributional shifts or retraining to new target tasks. We validate the effectiveness of \texttt{RESQUE} on CNNs and ViTs. \texttt{RESQUE} is shown to be an effective estimator when retraining to various distributional shifts. \texttt{RESQUE} is also utilized to estimate resources and performance when retraining to different target tasks. It is evaluated on different original tasks that are retrained to new target tasks. 
 All the results consistently validate that \texttt{RESQUE} is an effective estimator of various retraining measures, including energy consumption and carbon emission, enabling sustainable decisions with regard to model reusability.

\bibliography{aaai25}

\clearpage
\section{Appendix}

\subsection{Hardware and software setup}
All experiments were carried out on NVIDIA RTX GPUs, specifically utilizing the NVIDIA RTX 2060, 2070, 2080, 3060Ti, 4060Ti, and A100 models. For measuring carbon emissions and energy consumption, all experiments were run on RTX 2060 node. The codebase was developed and run on machines with Ubuntu 20.4 OS. Library and framework versions are submitted in the requirements file of the codebase.

\subsection{Hyperparameters}
The ResNet and VGG models were trained using the Adam optimizer, while the ViT model was trained with the SGD optimizer. For training the original models on the clean distribution, an initial learning rate of 0.001 was set for ResNet and VGG, and 0.01 for ViT, with the ViT learning rate being reduced by a factor of 10 after the 70th epoch. To mitigate overfitting, L2 regularization with a weight decay of 0.0001 was applied. Additionally, image augmentations, including random horizontal flip, random rotation, random affine transformation, color jitter, and normalization, were performed. When retraining the model to distribution shifted data or new task, the initial learning rate for models optimized using the Adam optimizer was set to 0.0001, and it was reduced by a factor of 10 after the 40th, 70th, and 90th epoch. A similar learning rate schedule was used for models optimized using SGD optimizer, but the initial learning rate for retraining was set to 0.001. For retraining to new distributions and new tasks, the L-2 regularization term was set to $1e-05$, and similar image augmentation schemes as clean training was utilized. 

For setting the hyperparameters, other values of learning rates starting from 0.1 to 0.0001 were explored, as well as L-2 regularization term with value 0.001, and different image augmentation magnitudes. We performed a single seed run with lowered target accuracy to find the most optimal learning rates and tuned it based on the actual target accuracy and training duration. For image augmentations, random horizontal flip with probability of 50\%, random rotation between 25\textdegree to 45\textdegree, random affine of 20\textdegree and color jitter value of 0.1 was set. All images are normalized using a mean and standard deviation of 0.5, across all channels.

\end{document}